\documentclass[conference]{IEEEtran}
\usepackage[noadjust]{cite}

\newcommand{\kETAL}    {{\em et al. }}
\usepackage{graphicx}
\usepackage{amsmath}
\usepackage{color,soul}
\usepackage{multirow}
\usepackage[ruled,vlined]{algorithm2e}
\SetKwRepeat{Do}{do}{while}%
\usepackage[nodisplayskipstretch]{setspace}
\makeatletter
\def\@copyrightspace{\relax}
\makeatother

\begin{document}
	
	\title{\huge Towards Budget-Driven Hardware Optimization for Deep Convolutional Neural Networks using Stochastic Computing \vspace{-0.4cm}} 

			
			
	\author{\IEEEauthorblockN{Zhe Li\IEEEauthorrefmark{1}, Ji Li\IEEEauthorrefmark{2}, Ao Ren\IEEEauthorrefmark{1}, Caiwen Ding\IEEEauthorrefmark{1},Jeffrey Draper\IEEEauthorrefmark{3}\IEEEauthorrefmark{2}, Qinru Qiu\IEEEauthorrefmark{1},Bo Yuan\IEEEauthorrefmark{4}, Yanzhi Wang\IEEEauthorrefmark{1}}
		\IEEEauthorblockA{\footnotesize\IEEEauthorrefmark{1}Department of Electrical Engineering and Computer Science, Syracuse University, Syracuse, NY 13244, USA\\
			}
		
		\IEEEauthorblockA{\footnotesize\IEEEauthorrefmark{2}Department of Electrical Engineering, University of Southern California, Los Angeles, CA 90089, USA\\
			}
		
		\IEEEauthorblockA{\footnotesize\IEEEauthorrefmark{3} Information Sciences  Institute, University of Southern California, Marina del Rey, CA 90292, USA\\ 
		}
		\IEEEauthorblockA{\footnotesize\IEEEauthorrefmark{4} Department of Electrical Engineering, City University of New York, City College, New York, NY 10031, USA\\ 
		\{zli89, aren, cading, qiqiu, ywang393\}@syr.edu, jli724@usc.edu, 	draper@isi.edu, byuan@ccny.cuny.edu  \vspace{-0.5cm}}
	}
	\maketitle
	\begin{abstract}
	Recently, Deep Convolutional Neural Network (DCNN) has achieved tremendous success in many machine learning applications.
	Nevertheless, the deep structure has brought significant increases in computation complexity. 
	Large-scale deep learning systems mainly operate in high-performance server clusters, thus restricting the application extensions to personal or mobile devices. 
	Previous works on GPU and/or FPGA acceleration for DCNNs show increasing speedup, but ignore other constraints, such as area, power, and energy. 
	Stochastic Computing (SC), as a unique data representation and processing technique, has the potential to enable the design of fully parallel and scalable hardware implementations of large-scale deep learning systems.
	This paper proposed an automatic design allocation algorithm driven by budget requirement considering overall accuracy performance.
	This systematic method enables the automatic design of a DCNN where all design parameters are jointly optimized.
	Experimental results demonstrate that proposed algorithm can achieve a joint optimization of all design parameters given the comprehensive budget of a DCNN.
	\end{abstract}
\section{Introduction}
\label{Sec:Introduction}

As the important branches of deep learning,~\textit{Deep Neural Networks (DNNs)} and Recurrent Neural Networks (RNNs), outperforms traditional machine learning techniques in several real-world problems~\cite{ding2017c, wang2018clstm,LinLNLDWP18}.
Recently, \textit{Deep Convolutional Neural Network (DCNN)}, which is one of most widely used types of DNNs, has achieved tremendous success in many machine learning applications, such as speech recognition \cite{sainath2013deep}, image classification \cite{simonyan2014very, WangD0YLMYQTQL18}, and video classification \cite{karpathy2014large}.
DCNN is now the dominant approach for almost all recognition and detection tasks, and approaches human performance on some tasks \cite{lecun2015deep}.
Nevertheless, the introduction of deep structure in deep learning brought significant increases in computation complexity.
Large-scale deep learning systems mainly operate in high-performance server clusters, thus restricting the application extensions to personal or mobile devices.

\textit{General-Purpose Graphics Processing Units (GPGPUs)} are widely used for current deep learning research to accelerate DCNNs \cite{ciresan2011flexible}.
GPGPU's major competitor is \textit{Field-Programmable Gate Arrays (FPGAs)}.
Considering energy-efficiency, FPGAs are more suitable for portable and embedded DCNN applications.
Previous works on FPGA acceleration for DCNNs \cite{farabet2009cnp}\cite{chakradhar2010dynamically}\cite{cadambi2010programmable} show increasing speedup.
These implementations, however, focus on improving the throughput of an embedded network, which ignores other constraints to run a network, such as area, power, and energy.
A notable trend is that machine learning is running locally on mobile/wearable devices and \textit{Internet-of-Things (IoT)} entities instead of relying on a remote server.
In order to bring the success of DCNNs to these resource constrained systems, designers must overcome the challenges of implementing resource-hungry DCNNs in embedded systems with limited area and power budget.

\textit{Stochastic Computing (SC)} \cite{gaines1969stochastic},
as a unique data representation and processing technique, has the potential to enable the design of fully parallel and scalable hardware implementations of large-scale deep learning systems.
Many complex arithmetic operations can be implemented with very simple hardware logic in stochastic computing framework \cite{gaines1969stochastic, li2017towards,li2017hardware,yuan2017softmax,li2017normalization}, which offers an immense design space for (i) neuron integrations due to the significantly reduced area per neuron, and (ii) performance optimizations with respect to the \textit{budget} of area, error resiliency, power/energy, or speed.
We use the term budget to describe the design constraint when implementing a hardware based DCNN using SC.
For example, given the area budget of an embedded design of DCNN, SC enables a comprehensive optimization including other design parameters mentioned above.

A node, referred as to \textit{neuron}, in a DCNN can be implemented by different stochastic computing based designs.
Given these designs, how to arrange them to structure a complete DCNN achieving preferred design parameter(s) such as constrained energy, promised accuracy, and restricted area leaves blanks to researchers.
This paper deals with the problem of deriving the optimal structure of hardware deep learning systems given a design budget and proposes an automatic algorithm driven by budget requirement with the comprehensive design parameters of a network taken into consideration.
More specifically, the proposed automatic design allocation algorithm greedily decides implementation for each layer of a DCNN, and then optimizes the complete DCNN jointly to achieve the better objective design parameter(s) by re-allocating different implementations.
This systematic method enables the automatic design of a DCNN where design parameters are jointly optimized given the budget.

The contributions of this paper are summarized as follows,
\textit{1)} SC paradigm is finely applied to DCNN, i.e., major computing tasks are performed in SC domain. With SC components, the hardware footprints can be significantly reduced for wearable/mobile devices.
\textit{2)}  We explored different implementations for neurons of different layers in a DCNN.
\textit{Accumulative/Approximate Parallel Counter (APC)}~\cite{parhami1995accumulative,kim2016dynamic} and \textit{Multiplexer} based neurons with distinct implementation details are investigated.
\textit{3)} An automatic design allocation algorithm is proposed to optimize the complete DCNN hardware design using SC.
This algorithm can also evaluate a general budget-driven hardware optimization for DCNNs.

Experimental results have been demonstrated on the problem of classifying handwritten digits in MNIST database using LeNet-5 \cite{lecun1995comparison} with the SC based DCNN designs. It reveals that proposed automatic design allocation algorithm can achieve a joint optimization of all design parameters given the comprehensive budget of a DCNN.


\section{Related Works}
\label{Sec:Related}

Experiments in \cite{ciresan2011flexible}\cite{strigl2010performance}
showed a high speedup of DCNN implemented on GPGPU. 
However, the widespread deployment of DCNNs has been hindered by their high complexities and power consumptions of server-based GPGPU-like devices, particularly in resource-constrained offline wearable devices and embedded systems.
Farabet \kETAL in \cite{farabet2009cnp} proposed an FPGA-based processor specific for convolutional networks.
The proposed processor was well structured. Nevertheless, the paper lacked evaluation of power and/or energy consumption.
Similarly, Cadambi \kETAL of \cite{cadambi2010programmable} demonstrated a general accelerator working for five typical tasks including DCNN. 
The paper, however, explored architectural design space instead of considering optimization of on-chip design parameters of a hardware-based DCNN.
In \cite{chakradhar2010dynamically}, authors contributed to developing a co-processor to improve the speed of DCNN, but they didn't mention how to optimize the network performance given hardware budget constraints.

Predecessors have been investigating SC as a candidate for implementing hardware-based DCNN.
Inspired by \cite{parhami1995accumulative}, authors in \cite{ting2014stochastic} introduced several basic hardware implementations for matrix operations including inner-product calculation which is crucial in DCNN inference process.
Based on lots of similar works exploring hardware implementation for stochastic computing, a recent work \cite{kim2016dynamic} presented a neuron cell design using SC components, where the progressive precision characteristics of SC was exploited.
The aforementioned papers focused on the analysis of performance in SC implementations, but optimization was not accomplished from a higher network-wise view. Even though the synthesis results were listed, but still, there lacked re-design of the implementation for DCNN with constrained hardware resources.
The aforementioned researches proved SC is a promising technique for embedded DCNN system. However, there is no existing work designing DCNN using SC with a structural optimization given hardware budget.

\section{Stochastic Computing Based DCNN}
\label{Sec:SC}

\subsection{Deep Convolutional Neural Network Architecture}
In this paper, we consider a general DCNN architecture, which consists of a stack of convolutional layers, pooling layers, and fully connected layers.
By arranging the topology of the above layers, powerful architectures, such as LeNet \cite{lecun1998gradient} and AlexNet \cite{krizhevsky2012imagenet}, can be built for specific applications.
Without the loss of generality, we conduct the investigation on the fifth generation of LeNet architecture using SC, which is comprised of two pairs of convolutional and pooling layers, and one fully connected hidden layer with the output layer. 
Note that the proposed methodology can accommodate other DCNN architectures as well.

A convolutional layer is associated with a set of learnable filters (or kernels), which are activated when specific types of features are found at some spatial positions in the inputs.
After obtaining features using convolution, a subsampling step can be applied to aggregate statistics of these features to reduce the dimensions of data and mitigate over-fitting issues.
This subsampling operation is realized by a pooling layer in hardware-based DCNNs, where different non-linear functions can be applied, such as max pooling, average pooling, and L2-norm pooling.
The activation functions in neurons are non-linear transformation functions, such as Rectified Linear Units (ReLU) $f(x)=max(0,x)$, hyperbolic tangent (tanh) $f(x)=tanh(x)$ or $f(x)=|tanh(x)|$, and sigmoid function $f(x)=\frac{1}{1+e^{-x}}$. We adopt the tanh activation function since it can be implemented efficiently as a finite state machine (FSM) in SC using a stochastic approximation method.
The fully connected layer is a normal neural network layer with its inputs fully connected with its previous layer.
The loss function of DCNN specifies how the network training penalizes the deviation between the predicted and true labels, and typical loss functions are softmax loss, sigmoid cross-entropy loss or Euclidean loss.

\begin{figure}
	\centering
	\includegraphics[width=0.85\columnwidth]{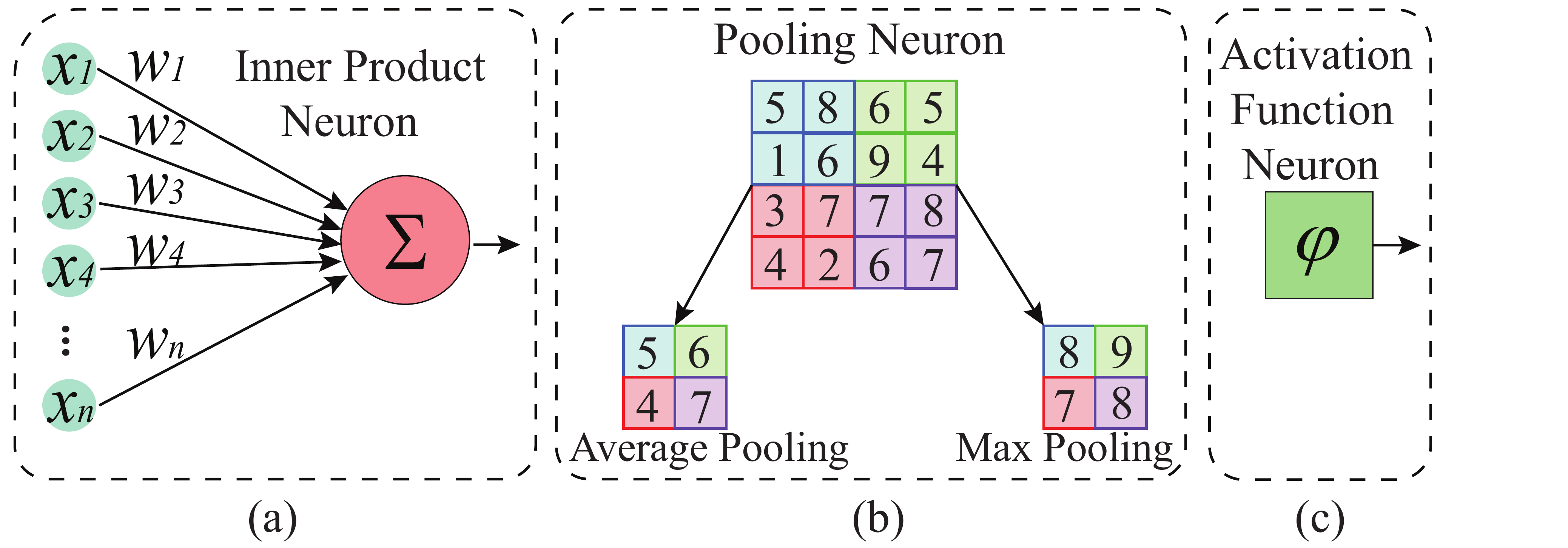}
	\vskip -0.8em
	\caption{Neurons in DCNN. (a) Inner Product, (b) pooling, and (c) activation }
	\label{fig:neuronconcept}
\end{figure}
In general, we can define three kinds of basic neurons (nodes) in hardware-based DCNN based on their corresponding operations as Fig.\ref{fig:neuronconcept} shows. Neurons in convolutional layers and fully connection layers calculate the inner product shown in Fig.\ref{fig:neuronconcept} (a) of inputs and weights based on its incoming connection with the previous layer.
And the products are subsampled through a pooling neuron shown in Fig.\ref{fig:neuronconcept} (b). We use average pooling here as a case study due to its simple hardware implementation.
The subsampled outputs are transformed by an activation function shown in Fig.\ref{fig:neuronconcept} (c) to ensure the inputs of next layer are within $[-1,1]$.

\subsection{Stochastic Computing Based Neuron}
In stochastic computing, the value of a number which lies in $[0,1]$ is represented by the occurrence probability of 1s in a random bit stream.
A $4$-bit sequence $X=0010$, for example, represents $x=P(X=1)=\frac{1}{4}=0.25$.
An $m$-bit sequence can only represent numbers in the set $\{\frac{0}{m},\frac{1}{m},\frac{2}{m},\cdots,\frac{m}{m}\}$; Only a small subset of the real numbers in the interval $[0,1]$ can be expressed exactly in SC.
Clearly, the precision and accuracy of SC depend on the length of the stream.

The two most popular representations for stochastic numbers are unipolar and bipolar formats, which interpret values in the intervals $[0, 1]$ and $[-1, 1]$, respectively.
Unipolar coding is commonly used in unsigned arithmetic operations, whereas bipolar format is used in signed arithmetic calculations.
More specifically, in unipolar coding, the number $x$ carried in a stochastic stream of bits $X$ is
$x=P(X=1)=P(X)$,
whereas in the bipolar format,
$x=2P(X=1)-1=2P(X)-1$.

In this section, we first conduct a detailed investigation of the energy-accuracy trade-off among two hardware neuron designs using SC, i.e., APC-based neuron and MUX-based neuron, as shown in Fig.\ref{fig:apc_neuron} and Fig.\ref{fig:mux_neuron}, respectively.
Hardware-based pooling is provided afterward, and finally, we present the structure optimization method for the overall DCNN architecture.

\subsubsection{APC-Based Neuron}\label{sec:apc_neuron}
\begin{figure}
	\centering
	\includegraphics[width=0.6\columnwidth]{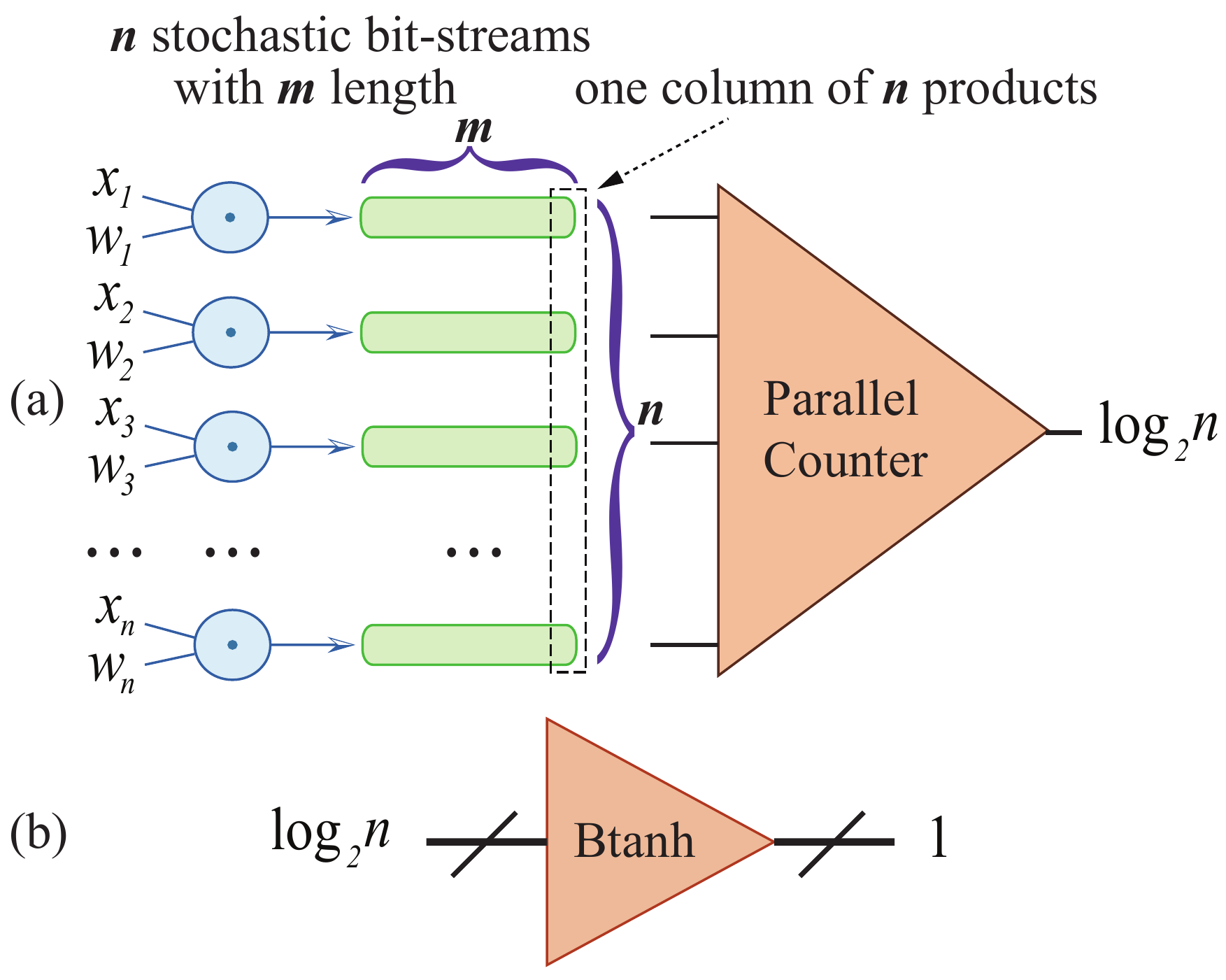}
	\vskip -0.8em
	\caption{APC-based neurons. (a) Inner product and (b) activation}
	\label{fig:apc_neuron}
\end{figure}

Fig.\ref{fig:apc_neuron} illustrates the APC-based hardware neuron design, where the inner product is calculated using XNOR gates (for multiplication in bipolar coding) \cite{brown2001stochastic} and an APC (for addition).
To be more specific, we denote the number of bipolar inputs and stochastic stream length by $n$ and $m$, respectively.
Accordingly, $n$ XNOR gates are used to generate $n$ products of inputs ($x_i's$) and weights ($w_i's$), and then the APC accumulates the sum of 1s in each column of the products shown in Fig.\ref{fig:apc_neuron} (a).
Instead of an FSM, a saturated up/down counter is used to perform the scaled hyperbolic tangent activation function $Btanh(\cdot)$ for binary inputs as Fig.\ref{fig:apc_neuron} (b) shows.
Details and optimization of the $Btanh(\cdot)$ activation function using a saturated up/down counter for binary inputs are demonstrated in \cite{kim2016dynamic}.

\subsubsection{MUX-Based Neuron}\label{sec:mux_neuron}
\begin{figure}[b]
	\centering
	\includegraphics[width=0.6\columnwidth]{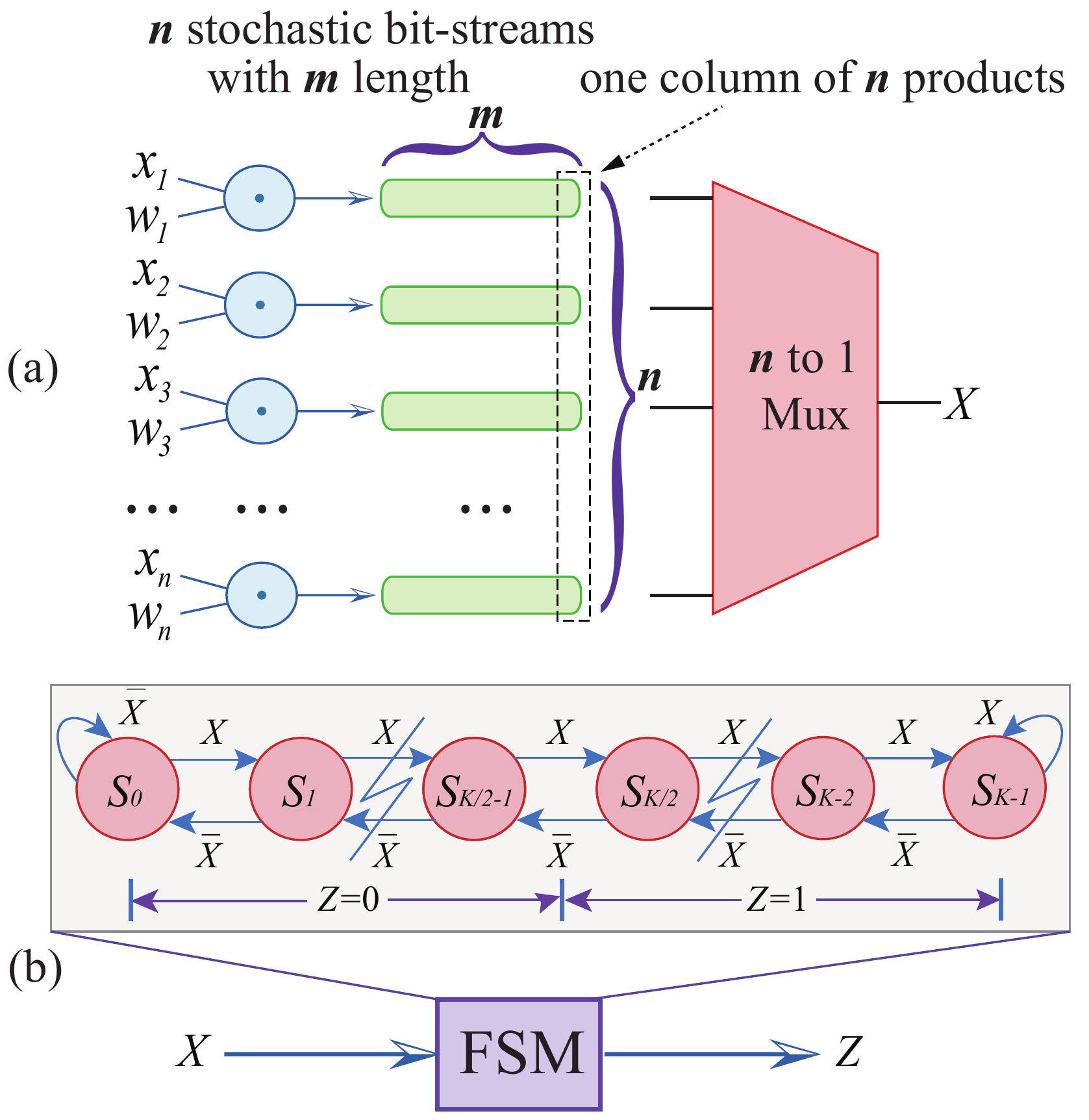}
	\vskip -0.8em
	\caption{MUX-based neurons. (a) Inner product and (b) activation}
	\label{fig:mux_neuron}
\end{figure}

As shown in Fig.\ref{fig:mux_neuron}, the MUX-based neuron is composed of XNOR gates, a MUX, and a $K$-state FSM.
XNOR gates compute the products of bipolar inputs ($x_i 's$) and weights ($w_i's$); the $n-$to$
-1$ MUX sums up all stochastic products; and the hyperbolic tangent activation function is achieved by a $K$-state FSM, respectively.
As the inner product calculated by a MUX is a stochastic number, the $K$-state FSM design mentioned in \cite{brown2001stochastic} can be used here to implement the activation function denoted as $Stanh(K,x) = tanh(\frac{K\cdot x}{2})$ where $x$ is the input.

Nevertheless, two problems challenge the implementation: (i) the inner product calculated by an $n$ input MUX is scaled to $\frac{z}{n}$, where the correct inner-product result is $z$, and (ii) with the input $\frac{z}{n}$, the $K$-state FSM calculates $tanh(\frac{K\cdot z}{2\cdot n})$ instead of the desired value $tanh(z)$.
Thus, in order to recover the correct activation, we need to re-scale up the results of MUX by $n$ times and multiply the stream by $\frac{2}{K}$ (or multiply by $\frac{2\cdot n}{K}$ directly).
As opposed to the relatively simple and efficient data conversions on a software platform, such conversions in a hardware-based neuron incur significant hardware overhead, because the linear gain transformation needs one more FSM \cite{brown2001stochastic}, and the scaling multiplication requires one XNOR gate as well as another bipolar stochastic stream generated.

In this paper, considering an $n$ inputs neuron with inner product denoted by $z$, we select the state number $K$ such that $\frac{2\cdot n}{K}=1$, and the final output of the FSM is calculated as
\begin{equation}
Stanh(K,\frac{z}{n})=tanh(\frac{K\cdot z}{2\cdot n})=tanh(z)\small
\end{equation}
In this way, we achieve the desired activation result with no additional bit stream conversion (i.e., no hardware overhead).


\subsubsection{Pooling Neuron} \label{sec_pooling}

In a DCNN, down-sampling steps are performed by the neurons in pooling layers, which reduce the dimensions of neurons for following convolutional layers or fully connection layers.
Pooling operation achieves the invariance to input data (i.e., image, video, etc.) transformations and better robustness to noise and clutter.
Moreover, the inter-layer connections can be significantly reduced for a hardware DCNN by using pooling layers.
In this paper, we adopt the \textit{average pooling} which simply outputs a mean of $k$ inputs.
In SC-based hardware, we implement it using a MUX and each input $x_i$ has the same probability to be selected as output.
For example, the stochastic arithmetic mean over 4 inputs ($2 \times 2$ region) is provided by using three 2-to-1 MUXs connected hierarchically, where selection signal for each MUX is a stochastic bit-stream for $0.5$.

\section{Budget-Driven Design Allocation}
\label{Sec:DesignAllocation}

Obviously, since neurons in each DCNN layer are homogeneous, we consider neurons in the same layer are implemented using the same design.
Design allocation which represents how to arrange different neuron designs for each DCNN's layer is a challenging task. Accuracy, power, area, and energy constraints have to be satisfied. 
Inspired by the \textit{stable marriage problem} \cite{gale2013college}, we present a design allocation algorithm which optimizes the overall network design given a preferred parameter.
The first step is to find a \textit{minimum feasible solution (MFS)} so that each layer in the DCNN is implemented by a valid neuron design. 
The second step is to optimize the MFS by re-allocating implementation for neurons in each layer to achieve the desired network-wise objective.
The objective of the overall DCNN optimization is evaluated in terms of a comprehensive design score defined as follows
\begin{equation}\small
	Score = \frac{\prod_j C_j^{\omega_j}}{1-Err}
\label{eqn:score}
\end{equation}
where $\omega_j$ is the integer weight of each design parameter, $C_j$ is the overall network cost in terms of one metric (e.g., area or power), and $Err$ is the overall network error rate.
As long as any one of the design parameter is the optimization target, corresponding $\omega_j$ increases.
A small score suggests an optimized design considering both design parameters and network performance.

\subsection{Minimum Feasible Solution}
The stable marriage algorithm cannot be applied directly to find a MFS.
In the stable marriage problem, an element can be matched to another element freely.
However, in our problem, a valid hardware design can implement multiple DCNN layers; several constraints result in limitation to apply a specific hardware design for neurons in a certain DCNN layer.
For example, in the case, where using APC-based neuron in a convolutional layer leads to the budget violation, the convolutional layer should not be implemented with the APC-based neuron.

\begin{algorithm}\scriptsize	
\DontPrintSemicolon
\KwData{list of feasible hardware implementations $\mathcal{I}$, list of layers in DCNN $\mathcal{L}$, optimization objective constraint $t$}
\KwResult{minimum feasible design allocation solution $\mathcal{S}$}
Initialize all $l \in \mathcal{L}$ to be free\;
$\mathcal{S} \gets \emptyset$\;
\While{$\exists$ free layer $l \in \mathcal{L}$ has no implementation} {
	\For{each $i \in \mathcal{I}$}{
		$IsValid \gets True$\;
			\If{cost $c_{lit} >$ Budget $b_{lt}$ }{
				$IsValid \gets False$\;
			}
		\If{IsValid}{
		$l$'s implementation $\gets i$ \;
		Append pair $(i,l)$ to $S$\;
		break\;
		}
	}
	\If{$l$ is free}{
		$\mathcal{S} \gets \emptyset$\;
		break\;
	}
}
Output $\mathcal{S}$
\caption{\small Minimum Feasible Design Allocation}
\label{Algo:DAMFS}
\end{algorithm}

In the minimum feasible design allocation algorithm shown in Algorithm.\ref{Algo:DAMFS}, the cost, denoted as $c_{lid}$, of an implementation $i$ on neurons in a layer $l$ with a specific constraint $d$ is defined as
\begin{equation}
c_{lid} = \psi \cdot u_{lid}
\label{eqn:cost}
\end{equation}
where $\psi$ is the number of neurons in the current layer and $u_{lid}$ is the unit cost of constraint $d$ for a neuron in layer $l$ by implementation $i$. 
Similarly, the budget for layer $l$ with respect to constraint $d$ is denoted by $b_{ld}$. 
Layers in a DCNN are firstly initialized to be free. 
The MFS algorithm tries to find a valid design for each layer at first. 
An implementation is \textit{valid} for a specific layer when the cost in every aspect of constraint metric satisfies the requirement.
The first valid implementation is used for current layer.
However, if all implementations are not valid, the solution is void. And an impossible solution is reported. 
This greedy process is repeated for each layer of a DCNN to obtain a MFS.

\subsection{Joint Design Allocation Optimization}
The MFS guarantees that each DCNN layer has a valid implementation design, but the complete DCNN is not optimized, i.e. the accuracy of DCNN is not optimized for the given constraints.
A joint optimization algorithm is proposed in Algorithm.\ref{Algo:DAOPT}.
The cost in the algorithm is defined in Eqn.(\ref{eqn:cost}), where we only calculate the cost for the constraint $t$. 
Taking the MFS and the given design budget as the inputs, the algorithm firstly seeks an implementation with lower cost for each DCNN layer. 
The cost difference between the old implementation and the new one has to be as big as $\theta$ to output the new one. 
After a lower cost design is ensured, the algorithm evaluates the refined implementation.
If the accuracy increases, which means the error rate of the network with new implementation is smaller than the previous one, the algorithm will output the new implementation scheme for the DCNN. 
When the process reaches an iteration limit $\tau$, the current solution is recognized as the optimized one and algorithm terminates.

\begin{algorithm}\scriptsize	
	\DontPrintSemicolon
	\KwData{minimum feasible design allocation solution $\mathcal{S}$, list of feasible hardware implementations $\mathcal{I}$, optimization objective constraint $t$}
	\KwResult{optimized design allocation solution $\mathcal{S'}$}
	$lastScore \gets Score_\mathcal{S}$\;
	$currScore \gets Score_\mathcal{S}$\;
	$iterCount \gets 0$\;
	\Do{$lastScore \le currScore$ or $iterCount < \tau$}{
		$lastScore \gets currScore$\;
		$lastCost \gets 0$\;
		$currCost \gets 0$\;	
		\Do{$lastCost - currCost \le $ threshold $\theta$}{
			$\mathcal{S'} \gets \emptyset$\;
			$lastCost \gets currCost$\;
			$currCost \gets 0$\;
			\For{each pair $(i,l)$ in $\mathcal{S}$}{
				\For{each $i' \in \mathcal{I}$}{
					\If{cost $c_{i't} < c_{it}$}
					{
						$l$'s implementation $\gets i'$\;
					}
				}
				$i^* \gets l$'s implementation\;
				$currCost \gets currCost + dc_{i^*}$\;
				Append pair $(i^*,l)$ to $\mathcal{S'}$\;
			}
		}
		$currScore \gets Score_\mathcal{S'}$\;
		$iterCount \gets iterCount +1$
	}
	Output $\mathcal{S'}$\;
	
	\caption{\small Design Allocation Optimization}
	\label{Algo:DAOPT}
\end{algorithm}
\section{Experimental Results}
\label{Sec:Evaluation}

\subsection{Comparison between APC-based and MUX-based neuron }
\begin{figure*}
	\centering
	\includegraphics[width=1.8\columnwidth]{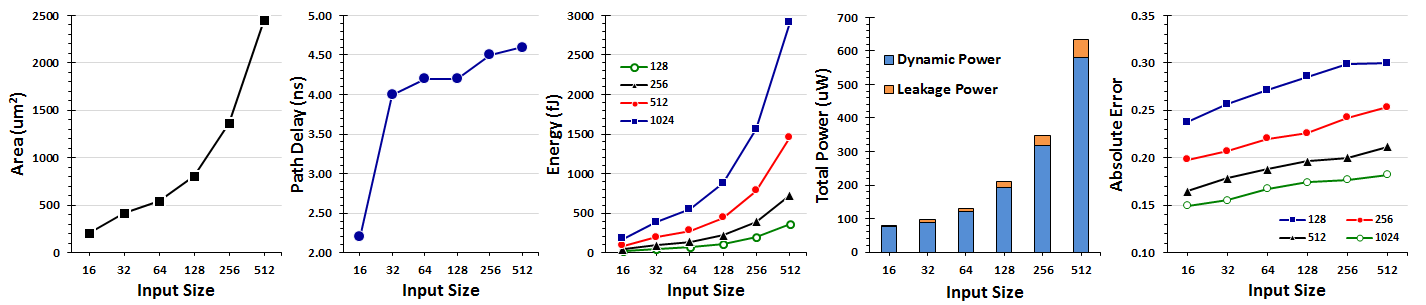}
	\vskip -0.8em
	\caption{Input size versus (a) area, (b) path delay, (c) energy, (d) power, and (e) absolute error with different bit-stream lengths for APC-based neuron}
	\label{fig:APC_results}
\end{figure*}
\begin{figure*}
	\centering
	\includegraphics[width=1.8\columnwidth]{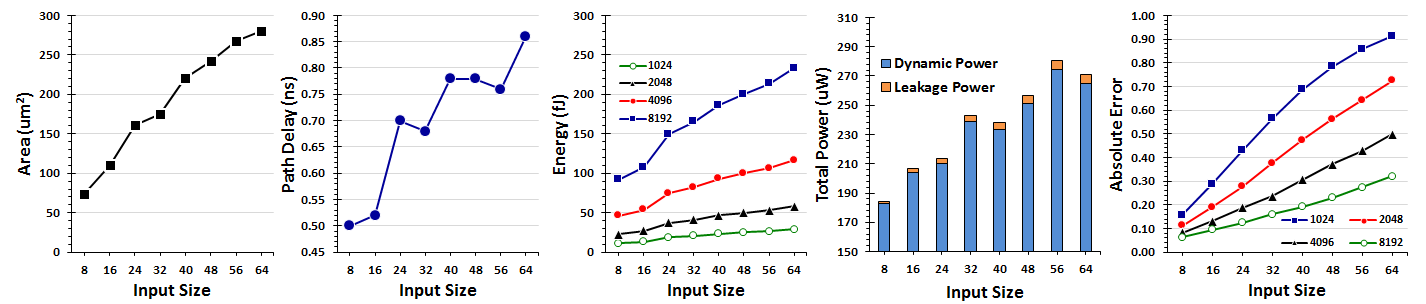}
	\vskip -0.8em
	\caption{Input size versus (a) area, (b) path delay, (c) energy, (d) power, and (e) absolute error with different bit-stream lengths for MUX-based neuron}
	\label{fig:MUX_results}
	\vspace{-1em}
\end{figure*}

We use Synopsys Design Compiler to synthesize the neurons with the 45nm Nangate Open Cell Library \cite{nangate}.
For an APC-based neuron, the area, path delay, energy, power, and absolute error with respect to the input size are shown in Fig.\ref{fig:APC_results} (a), (b), (c), (d), and (e), respectively. Energy and absolute error differ due to bit-stream length change, while other measures remain constant.
To be more specific, as illustrated in Fig.\ref{fig:APC_results} (a), (c), and (d), the APC-based neuron shows an exponential increase in area, energy, and power including dynamic and leakage power as the input size of a neuron increases exponentially.
This means that the area, energy, and power are linearly proportional to input size.
However, path delay shown in Fig.\ref{fig:APC_results} (b) reflects a saturated pattern when the input size of a neuron increases to a certain point, which means that a large input size will not lead to extreme long path delay.

The reason results from: With the efficient implementation of $Btanh(\cdot)$ function, the hardware of $Btanh(\cdot)$ increases logarithmically as the input increases, since the input width of $Btanh(\cdot)$ is $log_2 n$.
On the other hand, the number of XNOR gates and the size of the APC grow linearly as the input size increases.
Hence, the inner product calculation part, i.e., XNOR array and APC, is dominant in an APC-based neuron, and the area, power, and energy of the entire APC-based neuron cell also increase at the same rate as the inner product part when the input size increases.

We observe that the error is normally distributed with a mean value of $0$.
In this paper, we denote absolute error as the absolute standard deviation of error's distribution.
For different bit-stream lengths, the result, as Fig.\ref{fig:APC_results} (e) shows, agrees on the intuitive observation that a longer bit stream can reduce the absolute error of calculation.
Moreover, more inputs lead to larger absolute error.
The absolute error increases logarithmically with respect to input size.
The improvement due to the increase of bit-stream length is non-linear but independent of the input size.
Longer bit-stream length helps to reduce the error, but this improvement decreases and converges when bit-stream length gets longer than 1024.
Designers should consider the latency and energy overhead caused by long bit streams; the convergence of improvement trend helps designers achieve the desired trade-off between accuracy and overhead.

Similarly, we investigate the performance of the MUX-based neuron with respect to its input size.
Fig.\ref{fig:MUX_results} (a), (b), (c), and (d) show the results of the number of inputs versus the area, path delay, energy, power, and absolute error with respect to the input size.
Based on observation on the absolute error the MUX-based neuron can achieve, as shown in Fig.\ref{fig:MUX_results} (e), when the input size exceeds 64, the absolute errors with different bit-stream length approach 1, which is about 100\% error compared to correct value range.
Then we plotted the chart for input size from 8 to 64 only.
With input size increase linearly, the absolute error increase in a linear pattern.
The reason is that MUX addition selects only one bit at a time and ignores the rest of the bits, leading high absolute errors when input size is large.
Furthermore, as APC-based neuron performs, longer bit streams can reduce absolute error. The improvements are independent of input size; the improvements with respect to bit-stream length are logarithmic.
When the bit-stream length is increased large enough, the absolute error will not be reduced significantly.
In general, MUX-based neuron gains larger absolute error compared to APC-based neuron, which suggests MUX-based should be applied to more error-tolerant arithmetic operations.
In addition, one can observe from Fig.\ref{fig:MUX_results} (a), (c), and (d) that as the number of inputs increases, area, power, and energy of the MUX-based neuron all tend to increase.
The synthesis result also shows the path delay for a neuron increases approximate linearly when we enlarge the input size with the same stride.
These are because the MUX-based neuron with more inputs requires more XNOR gates and MUXes for inner product calculation, and more states in the FSM ($K=2\cdot n$) to compute the activation.
Hence, the increased hardware components result in more area, power, path delay, and energy in the neuron cell.

\begin{table}
	\centering
	\caption{Comparison between APC-based Neuron and MUX-based Neuron using $1024$ Bit Stream}\label{tbl1_cell_comparison}
	\resizebox{0.9\columnwidth}{!}{
		\begin{tabular}{|c|c|c|c|c|c|c|c|c|c|}
			\hline
			& \multicolumn{3}{c|}{\begin{tabular}[c]{@{}c@{}}APC-based  neuron\end{tabular}} & \multicolumn{3}{c|}{\begin{tabular}[c]{@{}c@{}}MUX-based  neuron\end{tabular}} \\ \hline
			Input size & 16     & 32     & 64     & 16     & 32     & 64       \\ \hline
			Absolute error       & 0.15   & 0.16   & 0.17   & 0.29   & 0.56   & 0.91    \\ \hline
			Area ($\mu m$\textsuperscript{2}) & 209.9  & 417.6  & 543.2  & 110.7  & 175.3  & 279.8  \\ \hline
			Path delay ($ns$) & 2.20  & 4.00  & 4.20  & 0.52  & 0.70  & 0.68  \\ \hline			
			Power ($\mu W$)       & 80.7   & 95.9   & 130.5  & 206.5  & 242.9  & 271.2   \\ \hline
			Energy ($fJ$)      & 177.4  & 383.7  & 548.1  & 110.0  & 169.1  & 238.9  \\ \hline
		\end{tabular}
	}
\end{table}
we compare the performance between APC-based neuron and MUX-based neuron using a fixed bit stream length equal to 1024 under different input sizes, as shown in Table.\ref{tbl1_cell_comparison}.
Clearly, APC-based neuron is more accurate but occupies more area than MUX-based neuron.
Besides, as APC is slower than MUX, the latency of APC-based neuron is larger than MUX-based neuron, which causes APC-based neuron to consume more energy than MUX-based neuron for one calculation.
As for the power performance, an APC-based neuron has less switching (due to the long latency) and larger area than the MUX-based neuron, resulting in less dynamic power, more leakage power, and less overall power.

\subsection{Evaluation of design allocation algorithm}
We use LeNet-5 DCNN as a case study in this experiment to evaluate our budget-driven algorithm to optimize a stochastic computing based DCNN.
Neurons in LeNet-5 DCNN layers are configured as 
$784(28\times28 )-11520(20\times24\times24)-2880(20\times24\times24)-3200(50\times8\times8)-800(50\times4\times4)-500-10$.
The MNIST handwritten digit image dataset \cite{deng2012mnist} consisting of 60,000 training data and 10,000 testing data with 28x28 grayscale image and 10 classes is used in the experiments.
The synthesis results are gathered as mentioned in Section \ref{sec:apc_neuron} using Synopsys Design Complier with the 45nm Nangate Open Cell Library \cite{nangate}.

We first listed several different configurations to implement each DCNN layer as Table.\ref{tbl2_conf} shows.
We fed these configurations into our algorithm given the budget constraint and evaluated the design score, defined as Eqn.(\ref{eqn:score}), for the optimized configuration of the DCNN.
To validate our algorithm, we optimized the DCNN with three different optimization targets as an example, i.e., area, power, and energy, given corresponding network-wise constraints.
The energy is approximately positively proportional to power; we set the weight for energy as 0 and use power to represent energy approximately.
In the experiment, three example cases were studied, where case 1 emphasized area and the rest emphasized power.

\begin{itemize}
\item Case1: $Score1 = \frac{Area^2 \cdot Power}{1-Err}$ and $Area \leq 5 mm^2$
\item Case2: $Score2 = \frac{Area \cdot Power^2}{1-Err}$ and $Power \leq 2 W$
\item Case3: $Score3 = \frac{Area \cdot Power^2}{1-Err}$ and $Energy \leq 4 \mu J$
\end{itemize}

As shown in Table.\ref{tbl3_score}, the proposed joint optimization algorithm picked configuration 4, 7, and 14 for case 1, 2, and 3, respectively. 
Under certain given constraints, some configurations are not valid, which is filtered out by the algorithm.
For comparison, we calculated the scores for those configurations which are not optimized when they are valid (configuration 1, 2, 9, 12).
We have conducted a separate exhaustive search, and verified the selected configurations by the proposed algorithm are the best.
One can observe from the Table.\ref{tbl3_score} that the picked configurations have the lowest score whereas other configurations are invalid or have larger scores. 
Also being observed from Table.\ref{tbl3_score}, our algorithm gives the best trade-off between network accuracy and design parameters (with the highest scores in all the cases). 


\begin{table}
	\centering
	\caption{Example Configurations}\label{tbl2_conf}
	\resizebox{0.8\columnwidth}{!}{
		\begin{tabular}{|c|c|c|c|c|}
			\hline
			Configuration & Bitstream Length & Layer 1 & Layer 2 & Layer 3 \\ \hline
1&	1024&	MUX&	MUX&	MUX \\ \hline
2&	1024&	MUX&	MUX&	APC \\ \hline
3&	1024&	MUX&	APC&	MUX \\ \hline
4&	1024&	MUX&	APC&	APC \\ \hline
5&	1024&	APC&	MUX&	MUX \\ \hline
6&	1024&	APC&	MUX&	APC \\ \hline
7&	1024&	APC&	APC&	MUX \\ \hline
8&	1024&	APC&	APC&	APC \\ \hline
9&	512&	APC&	MUX&	APC \\ \hline
10&	512&	APC&	APC&	MUX \\ \hline
11&	512&	APC&	APC&	APC \\ \hline
12&	512&	APC&	APC&	APC \\ \hline
\multicolumn{5}{|c|}{...} \\ \hline
13&	256&	APC&	APC&	MUX \\ \hline
14&	256&	APC&	APC&	APC \\ \hline
\multicolumn{5}{|c|}{...} \\\hline
			
		\end{tabular}
	}
\end{table}

\begin{table}
	\centering
	\caption{Examples of Configurations Explored and Generated by the Algorithm}\label{tbl3_score}
	\resizebox{\columnwidth}{!}{
		\begin{tabular}{|c|c|c|c|c||c|c|c|}
			\hline
			Configuration & Error (\%)& Area($mm^2$) & Power (W) & Energy (uJ) & Score 1 & Score 2 & score 3 \\ \hline
			1&	21.7 & 3.18	& 3.08	& 2.85 & 38.56 & -- & 38.56 \\ \hline			
			2&	11.9 & 3.69	& 3.03	& 4.21 & 38.50 & -- & -- \\ \hline
			4&	8.7 & 4.56  & 2.75	& 5.44 & \textbf{37.70} & -- & -- \\ \hline
			7&	4.3 & 7.20	& 1.77	& 7.63 & -- & \textbf{95.80} & -- \\ \hline			
			9&	4.7 & 6.83	& 2.01	& 3.96 & -- & -- & 28.84 \\\hline			
			12&	9.4 & 6.83	& 2.01	& 1.98 & -- & -- & 30.34 \\\hline		
			14&	2.0 & 7.70	& 1.72	& 2.36 & -- & 104.24 & \textbf{23.29} \\ \hline
\multicolumn{5}{|c||}{...} & \multicolumn{3}{c|}{...} \\ \hline

		\end{tabular}
	}
\end{table}

\section{Conclusion}
\label{Sec:Conclusion}

This paper introduced hardware implementation for Deep Convolutional Neural Network using stochastic computing.
Each distinct stochastic computing based neuron in DCNN is analyzed.
A two-step joint optimization algorithm is proposed that given design budgets, re-structuring the SC-based DCNN can achieve optimized hardware footprint with a relative high network accuracy performance.
Experimental results showed with restricted design requirements, the optimized SC-based implementation for DCNN achieved the lower error rate with the least design resources requested.

\scriptsize{
	\bibliographystyle{unsrt}
}
	\bibliography{reference}

\end{document}